\documentclass[preprint,10pt]{elsarticle}

\usepackage{amssymb}

\usepackage{listings}
\usepackage{subcaption}
\usepackage{hyperref}
\usepackage[scale=0.70]{geometry}
\usepackage{multicol}
\usepackage[cmex10]{amsmath}
\usepackage[ruled]{algorithm2e}
\usepackage{multirow}
\usepackage{paralist}
\usepackage{color}

\def \x {\mathbf{x}}

\def \w  {\mathbf{w}}
\definecolor{dkgreen}{rgb}{0,0.6,0}
\definecolor{gray}{rgb}{0.5,0.5,0.5}
\definecolor{mauve}{rgb}{0.58,0,0.82}

\def \x {\mathbf{x}}

\journal{Name of journal}

\begin{document}

\begin{frontmatter}



\title{SOL: A Library for Scalable Online Learning Algorithms\let\thefootnote\relax
}



%

\author{Yue Wu$^{\dag,\ddag}$, Steven C.H. Hoi$^\dag$, Chenghao Liu$^\dag$, Jing Lu$^\dag$, Doyen Sahoo$^\dag$, Nenghai Yu$^\ddag$\\
       $\dag$School of Information Systems, Singapore Management University, Singapore\\
       $\ddag$Department of EEIS, University of Science and Technology of China, China\\
       chhoi@smu.edu.sg, \{wye,ynh\}@mail.ustc.edu.cn, \{jing.lu.2014,doyen.2014\}@phdis.smu.edu.sg
}

\begin{abstract}
SOL is an open-source library for scalable online learning algorithms, and is particularly suitable for learning with high-dimensional data. The library provides a family of regular and sparse online learning algorithms for large-scale binary and multi-class classification tasks with high efficiency, scalability, portability, and extensibility. SOL was implemented in C++, and provided with a collection of easy-to-use command-line tools, python wrappers and library calls for users and developers, as well as comprehensive documents for both beginners and advanced users. SOL is not only a practical machine learning toolbox, but also a comprehensive experimental platform for online learning research. Experiments demonstrate that SOL is highly efficient and scalable for large-scale machine learning with high-dimensional data.
\end{abstract}


\end{frontmatter}

\setcounter{footnote}{0}

\section{Introduction}

In many big data applications, data is large not only in sample size, but also in feature/dimension size, e.g., web-scale text classification with millions of dimensions.
Traditional batch learning algorithms fall short in low efficiency and poor scalability, e.g., high memory consumption and expensive re-training cost for new training data. Online learning represents a family of efficient and scalable algorithms that sequentially learn one example at a time. Some existing toolbox, e.g., LIBOL \citep{hoi2014libol}, allows researchers in academia to benchmark different online learning algorithms, but it was not designed for practical developers to tackle online learning with large-scale high-dimensional data in industry.

In this work, we develop SOL as an easy-to-use scalable online learning toolbox for large-scale binary and multi-class classification tasks. It includes a family of ordinary and sparse online learning algorithms, and is highly efficient and scalable for processing high-dimensional data by using (i) parallel threads for both loading and learning the data, and (ii) specially designed data structure for high-dimensional data.
The library is implemented in standard C++ with the cross platform ability and there is no dependency on other libraries. To facilitate
developing new algorithms, the library is carefully designed and documented with high extensibility. We also provide python wrappers to facilitate experiments and library calls for advanced users.
The SOL website is host at \url{http://SOL.stevenhoi.org} and the software is made available \url{https://github.com/LIBOL/SOL}.

\section{Scalable Online Learning for Large-Scale Linear Classification}
\subsection{Overview}
Online learning operates sequentially to process one example at a time. Consider $\{(\x_t, y_t)|t\in[1,T]\}$ be a sequence of training data examples, where $\x_t \in R^d$ is a
$d$-dimensional vector, $y_t\in \{+1, -1\}$ for binary classification or $y_t \in \{0, \dots, C-1\}$ for multi-class classification ($C$ classes). As Algorithm~\ref{alg:framework} shows, at each time step $t$, the learner receives an incoming example $\x_t$ and then predicts its class label $\hat{y}_t$. Afterward, the true label $y_t$ is revealed and the learner suffers a loss $l_t(y_t, \hat{y}_t)$, e.g., the hinge loss is commonly used $l_t(y_t, \hat{y}_t) = max(0, 1 - y_t \cdot \hat{y}_t)$ for binary classification. For sparse online learning, one can modify the loss with $L1$ regularization $l_t(y_t, \hat{y}_t) = \hat{l}_t(y_t, \hat{y}_t) + \lambda \|\w_t\|_1$ to induce sparsity for the learned model $\w$. At the end of each learning step, the learner decides when and how to update the model.

\begin{algorithm}[ht]
	\footnotesize{
	\textbf{Initialize}: $\w_1=0$\;
	\For{$t$ in \{1,$\ldots$, T\}}{
	Receive  $\x_t \in R^d$, predict  $\hat{y}_t$, receive true label $y_t$\;
	Suffer loss $l_t(y_t, \hat{y}_t)$\;
	\If{$l_t(y_t, \hat{y}_t)$} {
	$\w_{t+1} \leftarrow update(\w_{t})$;
	}
	}
\caption{SOL: Online Learning Framework for Linear Classification}\label{alg:framework}
}
\end{algorithm}


The goal of our work is to implement most state-of-the-art online learning algorithms to facilitate research and application purposes on the real world large-scale high dimensional data. Especially, we include sparse online learning algorithms which can effectively learn important features from the high dimensional real world data~\citep{langford2009sparse}. We provide algorithms for both binary and multi-class problems.  These algorithms can also be classified into first order algorithms~\citep{xiao2010dual} and second order algorithms~\citep{crammer2009adaptive} from the model's perspective. The implemented algorithms are listed in table~\ref{tbl:algos}.

\begin{table}[htpb]
	\centering
	\scriptsize
    \begin{tabular}{c|c|c|l}
        \hline
        Type & {Methodology} & Algorithm & Description \\ \hline
		\multirow{11}{*}{\begin{minipage}{1.5cm}Online Learning\end{minipage}} & \multirow{5}{*}{First Order} &
			Perceptron~\citep{rosenblatt1958perceptron} &  The Perceptron Algorithm\\ \cline{3-4}
			&  & OGD~\citep{zinkevich2003online} &  Online Gradient Descent\\ \cline{3-4}
			&  & PA~\citep{crammer2006online} &  Passive Aggressive Algorithms\\ \cline{3-4}
			&  & ALMA~\citep{Gentile:2002:NAM:944790.944811} &  Approximate Large Margin Algorithm\\ \cline{3-4}
			&  & RDA~\citep{xiao2010dual} &  Regularized Dual Averaging\\ \cline{2-4}
		 & \multirow{6}{*}{Second Order} &
		 SOP~\citep{Cesa-Bianchi:2005:SPA:1055330.1055351} &  Second-Order Perceptron\\ \cline{3-4}
		 &  & CW~\citep{dredze2008confidence} &  Confidence Weighted Learning\\ \cline{3-4}
		 &  & ECCW~\citep{crammer2008exact} &  Exactly Convex Confidence Weighted Learning\\ \cline{3-4}
		 &  & AROW~\citep{crammer2009adaptive} &  Adaptive Regularized Online Learning\\ \cline{3-4}
		 &  & Ada-FOBOS~\citep{duchi2011adaptive} &  Adaptive Gradient Descent\\ \cline{3-4}
         &  & Ada-RDA~\citep{duchi2011adaptive} &  Adaptive Regularized Dual Averaging \\ \hline
		 \multirow{6}{*}{\begin{minipage}{1.5cm}Sparse\\Online
			 Learning\end{minipage}} &
			 \multirow{4}{*}{First Order} &
			 STG~\citep{langford2009sparse} &  Sparse Online Learning via Truncated Gradient\\ \cline{3-4}
			 &  & FOBOS-L1~\citep{duchi2009efficient} &  $l1$ Regularized Forward Backward Splitting\\ \cline{3-4}
			 &  & RDA-L1~\citep{xiao2010dual} &  Mixed $l1/l_2^2$ Regularized Dual Averaging\\ \cline{3-4}
         &  & ERDA-L1~\citep{xiao2010dual} &  Enhanced $l1/l_2^2$ Regularized Dual Averaging\\ \cline{2-4}
		 & \multirow{2}{*}{Second Order} & Ada-FOBOS-L1~\citep{duchi2011adaptive} &  Ada-FOBOS with $l1$ regularization\\ \cline{3-4}
         &  & Ada-RDA-L1~\citep{duchi2011adaptive} &  Ada-RDA with $l1$ regularization\\ \hline
    \end{tabular}
    \caption{Summary of the implemented online learning algorithms in SOL} \label{tbl:algos}
\end{table}

\subsection{The Software Package}

The SOL package includes a library, command-line tools, and python wrappers for the learning task. SOL is implemented in standard C++ to be
easily compiled and built in multiple platforms (Linux, Windows, MacOS, etc.) without dependency. It supports ``libsvm'' and ``csv'' data formats.
It also defined a binary format to significantly accelerate the training process. SOL is released under the Apache 2.0 open source license.

\subsubsection{Practical Usage}

To illustrate the training and testing procedure, we use the \emph{OGD} algorithm with a constant learning rate $1$ to learn a model
for ``\emph{rcv1}'' and save the model to ``\emph{rcv1.model}''.

\begin{lstlisting}
$ SOL_train --params eta=1 -a ogd rcv1_train rcv1.model
[output skipped]
$ SOL_test rcv1.model rcv1_test predict.txt
test accuracy: 0.9545
\end{lstlisting}

We can also use the python wrappers to train the same model.
The wrappers provide the cross validation ability which can be used to select
the best parameters as the following commands show. More advanced usages of SOL can be found in the documentation.

\begin{lstlisting}
$ SOL_train.py --cv eta=0.25:2:128 -a ogd rcv1_train rcv1.model
cross validation parameters: [('eta', 32.0)]
$ SOL_test.py rcv1.model rcv1_test predict.txt
test accuracy: 0.9744
\end{lstlisting}

\subsubsection{Documentation and Design}

The SOL package comes with detailed documentation. The README file gives an
``\emph{Installation}'' section for different platforms, and a ``\emph{Quick Start}'' section as
a basic tutorial to use the package for training and testing. We also provide
a manual for advanced users. Users who want to have a comprehensive evaluation
of online algorithms and parameter settings can refer to the ``\emph{Command
Line Tools}'' section. If users want to call the library in their own
project, they can refer to the ``\emph{Library Call}'' section. For those who want to
implement a new algorithm,  they can read the ``\emph{Design \& Extension of
the Library}'' section.
The whole package is designed for high efficiency, scalability, portability, and extensibility.
\begin{compactitem}
\item Efficiency: it is implemented in C++ and optimized to reduce time and memory cost.
\item Scalability: Data samples are stored in a sparse structure. All
	operations are optimized around the sparse data structure.
\item{Portability}:  All the codes follow the C++11 standard, and there is no dependency on external libraries. We use ``cmake'' to organize the project so that users on different platforms can build the library easily. SOL thus can run on almost every platform.
\item{Extensibility}: (i) the library is written in a modular way, including
	\emph{PARIO}(for PARallel IO), \emph{Loss}, and \emph{Model}. User can
	extend it by inheriting the base classes of these modules and implementing
	the corresponding interfaces; (ii) We try to relieve the pain of coding in
	C++ so that users can implement algorithms in a ``Matlab'' style. The code
	snippet in Figure 1 shows an example to implement the core function of the ``\emph{ALMA}'' algorithm.
\end{compactitem}

\subsection{Comparisons}

\begin{figure}
	\begin{minipage}[h][5.3cm][t]{.52\textwidth}
		\lstset{language=[GNU]C++,
		identifierstyle=\ttfamily,
		keywordstyle=\color[rgb]{0,0,1},
		commentstyle=\color{dkgreen},
		tabsize=2,
		frame=single
		}
		\begin{lstlisting}
		Vector<float> w; //weight vector
		void Iterate(SVector<float> x, int y) {
		  //predict label with dot product
		  float predict = dotmul(w, x);
		  float loss = max(0, 1 - y * predict); //hinge loss
		  if (loss > 0) { //non-zero loss, update the model
		    w = w + eta * y * x; //eta is the learning rate
		    //calculate the L2 norm of weight vector
		    float w_norm = Norm2(w);
		    if (w_norm > 1) w /= w_norm;
		  }
		}
		\end{lstlisting}
		\label{fig:code_alma}
		\vspace{-2mm}
		\caption{Example code to implement the core function of ``\emph{ALMA}'' algorithm.}
	\end{minipage}\hfill
	\begin{minipage}[h][5.3cm][t]{.45\textwidth}
		\centering
		\includegraphics[height=0.71\textwidth]{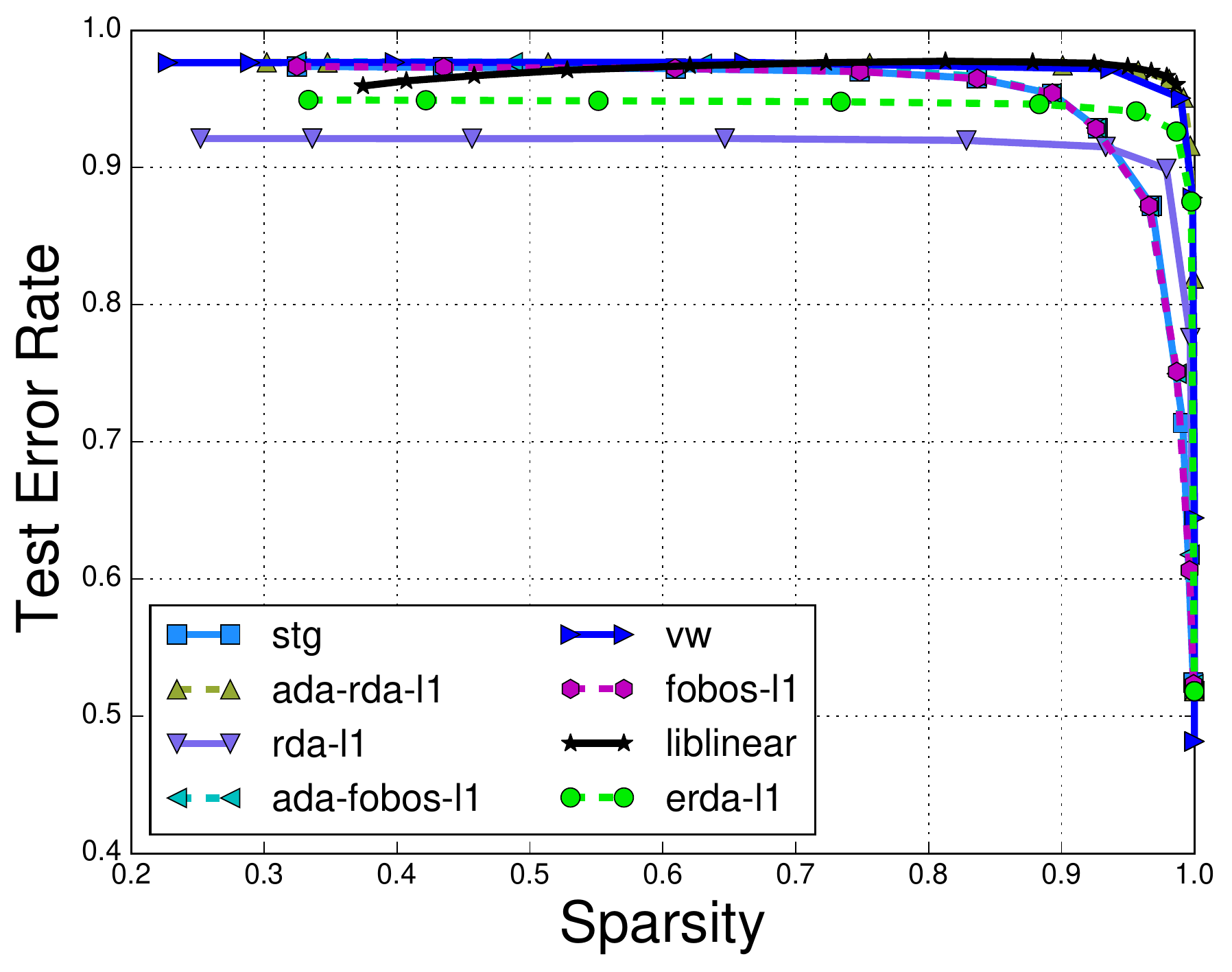}
		\vspace{-4mm}
		\caption{Comparison of Sparse Online Learning algorithms.}
		\label{fig:comp}
	\end{minipage}
\end{figure}

Due to space limitation, we only demonstrate that: 1) the online learning algorithms quickly reach comparable test accuracy compared to L2-SVM in
LIBLINEAR~\citep{Fan:2008:LLL:1390681.1442794} and VW \footnote{\url{https://github.com/JohnLangford/vowpal\_wabbit}. VW is another
OL tool with only a few algorithms}; 2) the sparse online learning methods can select meaningful features compared to L1-SVM in LIBLINEAR and L1-SGD in VW.
According to Table~\ref{tbl:ol_comp}, SOL provides a wide variety of algorithms that can achieve comparable test accuracies as LIBLINEAR and VW, while the training time is significantly less than LIBLINEAR. VW is also an efficient and effective online learning tool, but may not be a comprehensive platform for researchers due to its limited number of algorithms and somewhat complicate designs. Figure~\ref{fig:comp} shows how the test accuracy
varies with model sparsity. L1-SVM does not work well in low sparsity due to inappropriate regularization. According to the curves, the Ada-RDA-L1
algorithm achieves the best test accuracy for almost all model sparsity values. Clearly, SOL is a highly efficient and effective online learning toolbox.
More empirical results on other datasets can be found at \url{https://github.com/LIBOL/SOL/wiki/Example}.

\subsection{Illustrative Examples}
Illustrative examples of SOL can be found at: https://github.com/LIBOL/SOL/wiki/Example

\begin{table}
	\centering
	\footnotesize
    \begin{tabular}{|c|c|c|c|c|c|}
        \hline
		Algorithm & {Train Time(s)} & Accuracy & Algorithm & {Train Time(s)} & Accuracy \\ \hline
		Perceptron&  $8.4296\pm0.0867$& $0.9625\pm0.0014$ &  OGD & $8.4109\pm0.0982$ & $0.9727\pm0.0006$\\ \hline
		PA&  $8.4506\pm0.1031$& $0.9649\pm0.0015$ &  PA1 & $8.5113\pm0.1143$ & $0.9760\pm0.0005$\\ \hline
		PA2&  $8.4445\pm0.1068$& $0.9758\pm0.0003$ &  ALMA & $9.1464\pm0.1624$ & $0.9745\pm0.0009$\\ \hline
		RDA&  $8.4809\pm0.0899$& $0.9212\pm0.0000$ &  ERDA & $8.4623\pm0.1123$ & $0.9493\pm0.0002$\\ \hline
		CW&  $8.4356\pm0.1118$& $0.9656\pm0.0010$ &  ECCW & $8.4641\pm0.1116$ & $0.9681\pm0.0009$\\ \hline
	SOP & $8.5246\pm0.1017$ & $0.9627\pm0.0012$ &	AROW&  $8.4390\pm0.1292$& $0.9766\pm0.0002$ \\ \hline
	 Ada-FOBOS & $8.4897\pm0.0872$ & $0.9769\pm0.0003$	& Ada-RDA& $8.4388\pm0.1140$& $0.9767\pm0.0003$  \\ \hline
		VW&  $11.3581\pm0.3423$& $0.9754\pm0.0009$ &  LIBLINEAR & $77.9274\pm1.4742$ & $0.9771\pm0.0000$\\ \hline
    \end{tabular}
    \caption{Comparison of SOL with LIBLINEAR and VW on ``\emph{rcv1}''} \label{tbl:ol_comp}\vspace{-0.2in}
\end{table}

\section{Conclusion}

SOL is an easy-to-use open-source package of scalable online learning algorithms for large-scale online classification tasks. SOL enjoys high efficiency and efficacy in practice, particularly when dealing with high-dimensional data.
In the era of big data, SOL is not only a sharp knife for machine learning practioners in learning with massive high-dimensional data, but also a comprehensive research platform for online learning researchers.

\section*{Acknowledgements}
This work was done when the first author was an exchange student at Prof Hoi's research group.



\section*{References}

\begin{small}
\bibliographystyle{elsarticle-num}


\begin{thebibliography}{10}
\expandafter\ifx\csname url\endcsname\relax
  \def\url#1{\texttt{#1}}\fi
\expandafter\ifx\csname urlprefix\endcsname\relax\def\urlprefix{URL }\fi
\expandafter\ifx\csname href\endcsname\relax
  \def\href#1#2{#2} \def\path#1{#1}\fi

\bibitem{hoi2014libol}
S.~C. Hoi, J.~Wang, P.~Zhao, Libol: A library for online learning algorithms,
  The Journal of Machine Learning Research 15~(1) (2014) 495--499.

\bibitem{langford2009sparse}
J.~Langford, L.~Li, T.~Zhang, Sparse online learning via truncated gradient,
  The Journal of Machine Learning Research 10 (2009) 777--801.

\bibitem{xiao2010dual}
L.~Xiao, Dual averaging methods for regularized stochastic learning and online
  optimization, The Journal of Machine Learning Research 9999 (2010)
  2543--2596.

\bibitem{crammer2009adaptive}
K.~Crammer, A.~Kulesza, M.~Dredze, Adaptive regularization of weight vectors,
  Machine Learning (2009) 1--33.

\bibitem{rosenblatt1958perceptron}
F.~Rosenblatt, The perceptron: a probabilistic model for information storage
  and organization in the brain., Psychological review 65~(6) (1958) 386.

\bibitem{zinkevich2003online}
M.~Zinkevich, Online convex programming and generalized infinitesimal gradient
  ascent.

\bibitem{crammer2006online}
K.~Crammer, O.~Dekel, J.~Keshet, S.~Shalev-Shwartz, Y.~Singer, Online
  passive-aggressive algorithms, The Journal of Machine Learning Research 7
  (2006) 551--585.

\bibitem{Gentile:2002:NAM:944790.944811}
C.~Gentile, A new approximate maximal margin classification algorithm, J. Mach.
  Learn. Res. 2 (2002) 213--242.

\bibitem{Cesa-Bianchi:2005:SPA:1055330.1055351}
N.~Cesa-Bianchi, A.~Conconi, C.~Gentile, A second-order perceptron algorithm,
  SIAM J. Comput. 34~(3) (2005) 640--668.

\bibitem{dredze2008confidence}
M.~Dredze, K.~Crammer, F.~Pereira, Confidence-weighted linear classification,
  in: Proceedings of the 25th international conference on Machine learning,
  ACM, 2008, pp. 264--271.

\bibitem{crammer2008exact}
K.~Crammer, M.~Dredze, F.~Pereira, Exact convex confidence-weighted learning,
  in: Advances in Neural Information Processing Systems, 2008, pp. 345--352.

\bibitem{duchi2011adaptive}
J.~Duchi, E.~Hazan, Y.~Singer, Adaptive subgradient methods for online learning
  and stochastic optimization, Journal of Machine Learning Research 12 (2011)
  2121--2159.

\bibitem{duchi2009efficient}
J.~Duchi, Y.~Singer, Efficient online and batch learning using forward backward
  splitting, The Journal of Machine Learning Research 10 (2009) 2899--2934.

\bibitem{Fan:2008:LLL:1390681.1442794}
R.-E. Fan, K.-W. Chang, C.-J. Hsieh, X.-R. Wang, C.-J. Lin, Liblinear: A
  library for large linear classification, Journal of Machine Learning Research
  9 (2008) 1871--1874.

\end{thebibliography}
\end{small}



\clearpage

\section*{Required Metadata}
\label{}

\section*{Current executable software version}
\label{}

Ancillary data table required for sub version of the executable software: (x.1, x.2 etc.) kindly replace examples in right column with the correct information about your executables, and leave the left column as it is.

\begin{scriptsize}
\begin{table}[!h]
\begin{tabular}{|l|p{5.5cm}|p{5.5cm}|}
\hline
\textbf{Nr.} & \textbf{(executable) Software metadata description} & \textbf{Please fill in this column} \\
\hline
S1 & Current software version                           & v1.0.0 \\
\hline
S2 & Permanent link to executables of this version      & \url{https://github.com/LIBOL/SOL/archive/v1.0.0.zip} \\
\hline
S3 & Legal Software License                             & Apache 2.0 open source license\\
\hline
S4 & Computing platform / Operating System              & Linux, OS X, Windows. \\
\hline
S5 & Installation requirements \& dependencies          & Python 2.7 \\
\hline
S6 & Link to user manual                                & \url{https://github.com/LIBOL/SOL/wiki} \\
\hline
S7 & Support email for questions & chhoi@smu.edu.sg \\
\hline
\end{tabular}
\caption{Software metadata (optional)}
\label{}
\end{table}
\end{scriptsize}

\section*{Current code version}
\label{}

Ancillary data table required for subversion of the codebase. Kindly replace examples in right column with the correct information about your current code, and leave the left column as it is.

\begin{scriptsize}
\begin{table}[!h]
\begin{tabular}{|l|p{5.5cm}|p{5.5cm}|}
\hline
\textbf{Nr.} & \textbf{Code metadata description} & \textbf{Please fill in this column} \\
\hline
C1 & Current code version                                               & v1.0.0 \\
\hline
C2 & Permanent link to code/repository used of this code version        & \url{https://github.com/LIBOL/SOL/} \\
\hline
C3 & Legal Code License                                                 & Apache 2.0 open source license\\
\hline
C4 & Code versioning system used                                        & git \\
\hline
C5 & Software code languages, tools, and services used                  & Python/C/C++ \\
\hline
C6 & Compilation requirements, operating environments \& dependencies   & Python2.7/GCC/MSVC\\
\hline
C7 & If available Link to developer documentation/manual                & \url{https://github.com/LIBOL/SOL/wiki} \\
\hline
C8 & Support email for questions                                        & chhoi@smu.edu.sg \\
\hline
\end{tabular}
\caption{Code metadata (mandatory)}
\label{}
\end{table}
\end{scriptsize}
\end{document}